# MoDL-QSM: Model-based Deep Learning for Quantitative Susceptibility Mapping

[1]Ruimin Feng, [2]Jiayi Zhao, [3]He Wang, [3]Baofeng Yang, [1]Jie Feng, [1]Yuting Shi, [1]Ming Zhang, [4,5]Chunlei Liu, [6]Yuyao Zhang, [2]Jie Zhuang, [1,7]Hongjiang Wei

[1]School of Biomedical Engineering, Shanghai Jiao Tong University, Shanghai, China;

[2]School of Psychology, Shanghai University of Sport, Shanghai, China;

[3]Institute of Science and Technology for Brain-Inspired Intelligence, Fudan University, Shanghai, China;

[4]Department of Electrical Engineering and Computer Sciences, University of California, Berkeley, CA, USA;

[5]Helen Wills Neuroscience Institute, University of California, Berkeley, CA, USA.

[6]School of Information and Science and Technology, ShanghaiTech University, Shanghai, China

[7]Institute of Medical Robotics, Shanghai Jiao Tong University, Shanghai, China

**Correspondence Address:**
Hongjiang Wei, Ph.D.

School of Biomedical Engineering, Shanghai Jiao Tong University.

1954 Huashan Rd, MED-X Research Institute, Shanghai, 200030, China.

Phone: (021)6293 2050, Email: hongjiang.wei@sjtu.edu.cn

**Abstract**

Quantitative susceptibility mapping (QSM) has demonstrated great potential in quantifying tissue susceptibility in various brain diseases. However, the intrinsic ill-posed inverse problem relating the tissue phase to the underlying susceptibility distribution affects the accuracy for quantifying tissue susceptibility. Recently, deep learning has shown promising results to improve accuracy by reducing the streaking artifacts. However, there exists a mismatch between the observed phase and the theoretical forward phase estimated by the susceptibility label. In this study, we proposed a model-based deep learning architecture that followed the STI (susceptibility tensor imaging) physical model, referred to as MoDL-QSM. Specifically, MoDL-QSM accounts for the relationship between STI-derived phase contrast induced by the susceptibility tensor terms ($\chi_{13}$, $\chi_{23}$ and $\chi_{33}$) and the acquired single-orientation phase. The convolution neural networks are embedded into the physical model to learn a regularization term containing prior information. $\chi_{33}$ and phase induced by $\chi_{13}$ and $\chi_{23}$ terms were used as the labels for network training. Quantitative evaluation metrics (RSME, SSIM, and HFEN) were compared with recently developed deep learning QSM methods. The results showed that MoDL-QSM achieved superior performance, demonstrating its potential for future applications.

## 1. Introduction

Quantitative susceptibility mapping (QSM) is a recent technique based on MRI phase signal that quantifies the spatial distribution of magnetic susceptibility within a tissue (Acosta-Cabronero et al., 2016; Bilgic et al., 2012; Haacke et al., 2015; Li et al., 2016; Liu et al., 2015; Schweser et al., 2013; Shmueli et al., 2009; Wang and Liu, 2015; Wharton and Bowtell, 2010). QSM computes magnetic susceptibility from phase images of gradient-recalled echoes (GRE), typically assuming that phase shift results primarily from susceptibility-induced field inhomogeneity. The susceptibility contributors include biometals and molecules, e.g., iron, calcium, lipids, and myelin. Tissue susceptibility is also a crucial biomarker of pathological processes. For example, QSM has been applied for studying a variety of neurodegenerative diseases, e.g., brain aging, baby brain development, Parkinson's disease, Alzheimer's disease, and intracranial hemorrhage (Acosta-Cabronero et al., 2013; Barbosa et al., 2015; Betts et al., 2016; Chen et al., 2014; Du et al., 2018; Lotfipour et al., 2012; Zhang et al., 2019; Zhang et al., 2018). Furthermore, QSM has the potential to achieve accurate target localization in deep brain stimulation (DBS), benefitting from the improved visualization of subthalamic nuclei (STN) and globus pallidus pars internus (GPi) (Rasouli et al., 2018; Wei et al., 2019b).

Although QSM has demonstrated to have great potential for research purposes and clinical applications, QSM reconstruction is non-trivial. Computing susceptibility requires several processing steps involving phase unwrapping, background phase removal, and solving an inverse problem relating the tissue phase to the underlying susceptibility distribution. Among them, the dipole inversion for estimating the susceptibility map from a local tissue field map is more complicated. The field map must be deconvolved with a unit dipole kernel corresponding to a pointwise division in k-space. This deconvolution is ill-posed because of zeros in the k-space dipole kernel on two conical surfaces at approximately 54.7° relative to the $B_0$ direction. The inverse kernel is undefined at these surfaces and noise is amplified in regions where the kernel is very small, making a simple inversion of the forward calculation impractical. Different methods have been proposed to solve the ill-posed nature of this inverse problem. In general, QSM is solved by imposing conditions on the ill-posed inverse calculation to measure the susceptibility distribution while minimizing noise and artifacts (De Rochefort et al., 2008; Kressler et al., 2009; Liu et al., 2012; Wu et al., 2012). However, these regularization methods are considerably slow and care must be taken on the assumptions made when selecting spatial priors to avoid over-regularization and reduction of image contrast (Wharton and Bowtell, 2010).

Alternatively, another class of QSM algorithms uses multiple orientation sampling to compensate the missing data in a single orientation, such as Calculation of Susceptibility through Multiple Orientation Sampling (COSMOS) (Liu et al., 2009) and Susceptibility Tensor Imaging (STI) (Liu, 2010). These two methods require multiple scans for one subject with rotations of the head at different orientations relative to the main magnetic field. COSMOS requires at least three different scanning orientations and STI requires at least six to perform the dipole deconvolution analytically. Although both COSMOS and STI diagonal tensor element $\chi_{33}$ can be considered as the gold standard for different application purposes, the scanning time is significantly prolonged. On the one hand, this hinders their feasibility in clinical practice and increases the risk of motion artifacts, especially for the elderly and young children. On the other hand, large-angle rotations for human heads in a standard tunnel magnet with multi-channel head coils are limited. Thus, the multiple orientation QSM reconstruction methods are usually impractical for clinical studies.

Recently, convolutional neural networks (CNN) have been proposed to approximate the dipole inversion process and generate high-quality QSM from single orientation phase measurements. Some of these schemes adopted the U-Net structure (Ronneberger et al., 2015) as the backbone. For example, DeepQSM trained the network by synthetic geometric images to learn the relationship between phase and susceptibility (Bollmann et al., 2019). QSMnet used realistically acquired data to learn the COSMOS-like QSM maps (Yoon et al., 2018). QSMnet+ further improved the performance of QSMnet by data augmentation that enlarged the dynamic range of the training data to overcome underestimated high susceptibility values in brain nuclei (Jung et al., 2020). AutoQSM aimed to predict QSM directly from total phase maps without brain masking and background phase removal (Wei et al., 2019a). Inspired by the success of Generative Adversarial Networks (GANs) (Goodfellow et al., 2014) in computer vision, QSMGAN adopted GANs to further improve reconstruction quality and accuracy (Chen et al., 2020b). Other studies of QSM reconstruction included xQSM, which trained an octave CNN to improve network generalization capability (Gao et al., 2020). All of the above deep learning methods learn the necessary features of QSM maps via a data-driven manner. More recently, researchers proposed to incorporate the physical models into CNNs to solve the dipole inverse problem. For example, Variational regularizer for Nonlinear Dipole Inversion (VaNDI) trained a variational network (Hammernik et al., 2017) to optimize the parameters in an unrolled gradient descent algorithm for non-linear dipole inversion (Polak et al., 2019). Learned Proximal CNN (LPCNN) (Lai et al., 2020) and Proximal variational networks (Kames et al., 2019) combined the strengths of CNN and model-based iterative solvers to learn an implicit regularizer via proximity operator.

In this study, we proposed an STI-based deep learning architecture for single-orientation QSM reconstruction, referred to as MoDL-QSM. The main novelties of the proposed approach over related deep-learning schemes are (1) the proposed scheme accounts for the relationship between STI-derived phase induced by the rightmost column of the susceptibility tensor ($\chi_{13}$, $\chi_{23}$ and $\chi_{33}$) and the acquired single-orientation phase. (2) STI component, $\chi_{33}$, was used as the training label to make the network preserve the nature of anisotropic magnetic susceptibility in brain white matter. (3) We combined the STI physical model with CNNs to learn a regularizer implicitly. (4) MoDL-QSM can simultaneously learn $\chi_{33}$ and the field induced by $\chi_{13}$ and $\chi_{23}$ terms. Our results demonstrate that $\chi_{33}$ as network label makes the model-based framework more consistent with the physical assumption of single-orientation QSM algorithms. Only when $\chi_{13}$ and $\chi_{23}$ induced phase contribution is added to the physical model, the single orientation acquired phase can be appropriately interpreted by the learned susceptibility distribution. Additionally, MoDL-QSM can provide superior image quality and high accuracy susceptibility quantification for healthy and diseased brain tissues.

## 2. Theory

2.1 *Relationship Between Field Perturbation and Magnetic Susceptibility*

When biological tissue with susceptibility distribution $\chi$ is placed in the main magnetic field $B_0$, it will be magnetized and generate a magnetic field perturbation $\delta B$. Based on STI theory (Liu, 2010), $\chi$ is orientation-dependent, which can be represented by a second-order symmetric tensor, i.e., $\chi = \begin{bmatrix} \chi_{11} & \chi_{12} & \chi_{13} \\ \chi_{12} & \chi_{22} & \chi_{23} \\ \chi_{13} & \chi_{23} & \chi_{33} \end{bmatrix}$. In the subject frame of reference, the field perturbation $\delta B$ is given by:

$$\mathcal{F}\delta B^{sub} = \frac{\mathcal{F}\varphi}{2\pi\gamma T_E B_0} = \frac{1}{3}(\widehat{H}^{sub})^T \mathcal{F}\chi^{sub}\widehat{H}^{sub} - (k^{sub})^T\widehat{H}^{sub}\frac{(k^{sub})^T \mathcal{F}\chi^{sub}\widehat{H}^{sub}}{\|k^{sub}\|_2^2} \quad (1)$$

Where all superscripts in the formula indicate the coordinate system used. Standard notations indicate vector and scalar. Bold notations indicate matrix. $\mathcal{F}$ denotes the Fourier transform. $\gamma$ is the gyromagnetic ratio, $\varphi$ denotes tissue phase after background phase removal and $T_E$ is the echo time. $\widehat{H}^{sub} = [H_1^{sub}, H_2^{sub}, H_3^{sub}]^T$ is the unit direction vector of the applied main magnetic field. $k^{sub} = [k_x^{sub}, k_y^{sub}, k_z^{sub}]^T$ represents a spatial vector in the Fourier domain. $(\cdot)^T$ denotes transposition and $\|\cdot\|_2$ denotes the L2 norm. When the subject frame of reference is used, $\chi^{sub}$ remains the same for each orientation and the tensor can be solved by least-squares estimation after registering each orientation to the supine one (the normal position). When the laboratory frame of reference is used, the susceptibility tensor is rotated according to the rotation matrix for different head orientations, while the magnetic field vector remains along the z-axis, i.e., $\widehat{H}^{lab} = [0, 0, 1]^T$:

$$\chi^{lab} = \mathbf{R}^{\mathrm{T}} \chi^{sub} \mathbf{R} \tag{2}$$

Where **R** denotes the rotation matrix from the laboratory frame to the subject frame, which can be obtained by the registration process. For the single orientation GRE data acquisition, the imaging frame coincides with the laboratory frame. Considering the field is a function of $\boldsymbol{\mathcal{F}}\chi\widehat{H}^{lab}$, only the rightmost column of the tensor, $[\chi_{13}, \chi_{23}, \chi_{33}]^{\mathrm{T}}$, has contributions to field perturbation. Thus, the relationship between field perturbation and susceptibility tensor in the laboratory frame of reference is simplified as:

$$\boldsymbol{\mathcal{F}}\delta B^{lab} = \left[\frac{1}{3} - \frac{(k_z^{lab})^2}{\|\mathrm{k}^{lab}\|_2^2}\right]\boldsymbol{\mathcal{F}}\chi_{33}^{lab} - \frac{k_z^{lab}}{\|\mathrm{k}^{lab}\|_2^2}\left(k_x^{lab}\boldsymbol{\mathcal{F}}\chi_{13}^{lab} + k_y^{lab}\boldsymbol{\mathcal{F}}\chi_{23}^{lab}\right) \tag{3}$$

Where $\chi_{13}^{lab}, \chi_{23}^{lab}, \chi_{33}^{lab}$ are three components of the rightmost column of $\boldsymbol{\chi}^{lab}$. The field perturbation, $(\delta B')^{lab}$, derived from the off-diagonal tensor terms, $\chi_{13}^{lab}$ and $\chi_{23}^{lab}$, can be expressed as:

$$\boldsymbol{\mathcal{F}}(\delta B')^{lab} = -\frac{k_z^{lab}}{\|\mathrm{k}^{lab}\|_2^2}\left(k_x^{lab}\boldsymbol{\mathcal{F}}\chi_{13}^{lab} + k_y^{lab}\boldsymbol{\mathcal{F}}\chi_{23}^{lab}\right) \tag{4}$$

Conventional single-orientation QSM algorithms assume that the phase term, $(\delta B')^{lab}$, could be negligible and mainly focus on the susceptibility components along the $B_0$ direction, i.e., $\chi_{33}^{lab}$ as the ground truth susceptibility. Therefore, they aim to solve the following dipole inversion problem:

$$\boldsymbol{\mathcal{F}}\delta B^{lab} = \left[\frac{1}{3} - \frac{(k_z^{lab})^2}{\|\mathrm{k}^{lab}\|_2^2}\right]\boldsymbol{\mathcal{F}}\chi_{33}^{lab} \tag{5}$$

Eq. (5) motivates the use of $\chi_{33}^{lab}$ as the label to preserve the inherent susceptibility anisotropy of brain white matter. Additionally, the contributions from $\chi_{13}^{lab}$ and $\chi_{23}^{lab}$ components to the acquired single-orientation phase signal are nonnegligible since they could rise to 70% amplitude relative to $\chi_{33}^{lab}$ component (Langkammer et al., 2018). Therefore, we take $\chi_{33}^{lab}$ and $\delta B'$ as the learning targets of MoDL-QSM to ensure that the overall framework consistent with the STI model.

*2.2 Combination of QSM Physical Model and CNN*

Taking both $\chi_{33}^{lab}$ and $\delta B'$ into account, Eq. (3) can be rewritten as the following

linear equation system:

$$\delta B = \mathcal{F}^{-1}\left[\frac{1}{3} - \frac{(k_z)^2}{\|k\|_2^2} \quad 1\right]\mathcal{F}\begin{bmatrix}\chi_{33}\\ \delta B'\end{bmatrix} \tag{6}$$

Note that, for simplicity, we dropped the superscript "*lab*". $\mathcal{F}^{-1}$ is the inverse Fourier transform. Let $\boldsymbol{A}$ denote the forward operator matrix and vector $X$ denotes the voxel values of $\chi_{33}$ and $\delta B'$ to be solved.

$$\boldsymbol{A} = \mathcal{F}^{-1}\left[\frac{1}{3} - \frac{(k_z)^2}{\|k\|_2^2} \quad 1\right]\mathcal{F} \tag{7}$$

$$X = \begin{bmatrix}\chi_{33}\\ \delta B'\end{bmatrix} \tag{8}$$

Then Eq. (6) can be expressed as:

$$\delta B = \boldsymbol{A}X \tag{9}$$

Solving $X$ from $\delta B$ is an ill-posed problem. One solution is to convert Eq. (9) into a minimum optimization problem:

$$\hat{X} = \underset{X}{\mathrm{argmin}}\, g(X) + R(X) \tag{10}$$

Where $g(X) = \frac{1}{2}\|\boldsymbol{A}X - \delta B\|_2^2$ is the data consistency term. $R(X)$ is the regularization term added according to prior information. Generally, $R(X)$ is L1 norm containing non-differential points. One can use the proximal gradient descent algorithm (Parikh and Boyd, 2014) to iteratively solve Eq. (10). The $X$ in the $k^{th}$ iteration can be expressed as:

$$\hat{X}^k = Prox_{t_k,R}\left(\hat{X}^{k-1} - t_k\nabla g(\hat{X}^{k-1})\right) \tag{11}$$

and

$$\nabla g(\hat{X}^{k-1}) = \boldsymbol{A}^{\mathrm{H}}(\boldsymbol{A}\hat{X}^{k-1} - \delta B) \tag{12}$$

we have

$$\hat{X}^k = Prox_{t_k,R}\left(t_k\boldsymbol{A}^{\mathrm{H}}\delta B + (\boldsymbol{I} - t_k\boldsymbol{A}^{\mathrm{H}}\boldsymbol{A})\hat{X}^{k-1}\right) \tag{13}$$

Where $Prox_{t_k,R}(\cdot)$ is the proximity operator:

$$Prox_{t_k,R}(z) = \underset{X}{\mathrm{argmin}}\left(R(z) + \frac{1}{2t_k}\|X - z\|_2^2\right) \tag{14}$$

In the above formulas, $t_k$ is the step size in the $k^{th}$ iteration in gradient descent. $\boldsymbol{A}^{\mathrm{H}}$ is the conjugate transpose of $\boldsymbol{A}$. $\boldsymbol{I}$ represents the identity matrix. In Eq. (13), the operator $Prox_{t_k,R}(\cdot)$ does not depend on $g(X)$, only depends on $R(X)$, indicating that the data consistency term $g(X)$ and regularization term $R(X)$ can be decoupled during the solving process. Benefitting from this advantage of proximal gradient descent algorithm, we can incorporate CNNs into Eq. (13) to train a regularization term by learning its associated proximity operator, $Prox_{t_k,R}$. The scheme of the proposed architecture is shown in Fig. 1(a). We unrolled Eq. (13) into three iterations and initialized $\hat{X}^0=0$, so the input of the network was $t_0\boldsymbol{A}^{\mathrm{H}}\delta B$. $t_k$ in the $k^{th}$ iteration was a learnable parameter. Three CNNs shared their weights (blue blocks in Fig. 1(a), "**C**" represents "CNN") and replaced $Prox_{t_k,R}$ with the learnable parameters. The output of each CNN performed the physical model operation (green block in Fig. 1(a), "**P**" represents "Physical"). The final output of MoDL-QSM consisted of two channels, $\chi_{33}$ and $\delta B'$. The regularization term was learned by minimizing the L1 loss of the two channels separately between output and label.

## 3. Methods

*3.1 Network Architecture*

The network architecture is depicted in Fig. 1(b). A CNN has a total of 18 convolutional layers containing eight repetitive applications of ResBlock (He et al., 2016). For the first 17 convolutional layers, the kernel size is $3\times3\times3$ with stride 1. The batch normalization layer (BN) is used to speed up convergence. Rectified linear unit (ReLU) is adopted as the activation function. The last convolution is $1\times1\times1$ with 2 output channels to simultaneously generate $\chi_{33}$ and $\delta B'$ maps. The number of channels after each layer is summarized at the bottom of blocks and the output size is summarized at the top of blocks in Fig. 1(b).

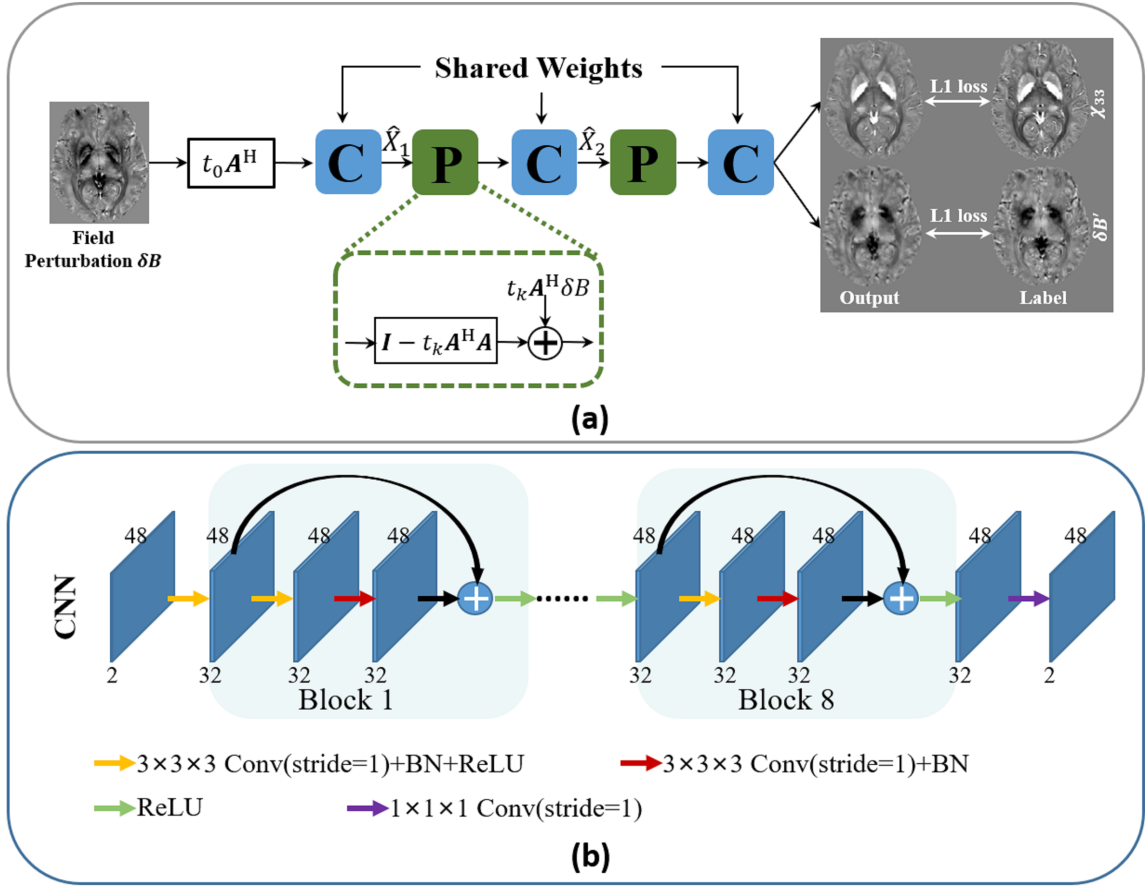

Figure 1. Overview of the proposed MoDL-QSM architecture. (a) The schematic diagram of MoDL-QSM. Three convolutional neural networks shared their weights (blue blocks in Fig. 1(a), "**C**" represents "CNN") and replaced $Prox_{t_k,R}$ with the learnable parameters. The output of each CNN performed the physical model operation (green block in Fig. 1(a), "**P**" represents "Physical"). The final output of MoDL-QSM consisted of two channels, $\chi_{33}$ and $\delta B'$. (b) The network structure of the CNN. The number of channels after each layer is summarized at the bottom of blocks and the output size is summarized at the top of blocks

## 3.2 MRI Data Acquisition and Processing

Our training data included 5 healthy volunteers with 15~23 head orientations per subject (see Table s1 in the supplemental material for detailed head rotation degrees). The subjects were scanned at Shanghai University of Sport using a 3T scanner (Prisma, Siemens Healthcare, Erlangen, Germany) equipped with a 64-channel head coil. This experiment was approved by Shanghai Jiao Tong University Human Ethics Committee and all subjects signed informed consent before scanning. A multi-echo 3D GRE sequence was used with the following scan parameters: FOV=210×224×160 mm$^3$, voxel size=1 mm$^3$ isotropic, TR=38 ms, TE$_1$/spacing/TE$_6$=7.7/5/32.7 ms, bandwidth=190

Hz/pixel, flip angle=15°, GRAPPA factor=2, and total imaging time=8.8 minutes. For each volunteer, the scan was repeated at different head orientations.

The brain mask was generated using BET in FSL (Smith, 2002) on the magnitude images. Phase images were processed in STISuite (https://people.eecs.berkeley.toolbox.edu/~chunlei.liu/software.html). First, the raw phase data was unwrapped by Laplacian-based phase unwrapping (Schofield and Zhu, 2003). Then the background phase was removed using VSHARP (Wu et al., 2012) to obtain the tissue phase. Additionally, the tissue phase from different echoes was normalized by $2\pi\gamma T_E B_0$ and averaged together to obtain the field map. Finally, the multiple orientation field maps were processed to compute the susceptibility tensor. Specifically, the magnitude images at the first $T_E$ of different head orientations were coregistered to that of the supine position by affine transformation with 7 degrees of freedom using FSL FLIRT (Jenkinson and Smith). The transformation matrix was then applied to the corresponding field perturbation map. The orientation of the applied magnetic field $\hat{H}^{sub}$ in the subject frame of reference (Eq. (1)) was calculated based on the transformation matrix. The six independent variables of the symmetric susceptibility tensor are computed voxel by voxel following the STI model (Liu, 2010). The obtained susceptibility tensor was converted to the laboratory frame of reference for each orientation according to Eq. (2). The field perturbation, $\delta B'$, induced by $\chi_{13}$ and $\chi_{23}$, was calculated based on Eq. (4). In total, 90 pairs of input field perturbation maps and the corresponding labels were obtained.

*3.3 Network Implementation*

The proposed network structure was implemented using Python 3.6.2 and Keras v2.2.5 with Tensorflow as the backend and trained using one NVIDIA 1080TI GPU. To fit into GPU memory, the patch size for MoDL-QSM training was cropped to 48×48×48. The training patches were generated by randomly extracting from all 90 scans. To improve training efficiency, patches with more than 20% background region were discarded. Twenty-seven thousand patches were obtained in total. 80% of them were used for training and 20% for validation. MoDL-QSM was trained for 40 epochs. The initial learning rate was $10^{-3}$ and reduced to half when the validation loss didn't decay in three epochs. The Adam optimizer (Kingma and Ba, 2015) was used with a batch of size 2. The total training time was about 30 hours. During the training process, 48×48×48 patches and the corresponding full-size dipole kernel ($\frac{1}{3} - \frac{(k_z)^2}{\|k\|_2^2}$) were fed into the network. To preserve high-frequency information, the 48×48×48 patches were padded to original size to match the size of dipole kernel when performing physical model operator (green blocks in Fig. 1(a)) and then were cropped back to 48×48×48.

For testing, MoDL-QSM can operate on full-size input data to save reconstruction time. If the GPU memory is not enough for the large size, a "patch-then-stitch" manner will be adopted. Specifically, the full-size images were cropped into 48×48×48 with 1/3 overlap. Then the output patches were stitched to produce complete output maps. The source codes and trained model ready for testing have been published at https://github.com/Ruimin-Feng/MoDL-QSM

*3.4 Evaluation of MoDL-QSM*

To demonstrate the proposed deep learning architecture with two labels is more rational than COSMOS as the training label, we conducted Experiment 1 to demonstrate the proposed method can well preserve susceptibility anisotropy of brain white matter.

**Experiment 1:** The original LPCNN architecture (Lai et al., 2020) was trained using the same 90 training data but with COSMOS as the label. Then the COSMOS labeled LPCNN and $\chi_{33}$ labeled MoDL-QSM were tested on another two subjects scanned at different orientations using the same scan parameters as the training data.

Then, to evaluate the reconstruction performance of MoDL-QSM, different datasets were used. Since these datasets differ from the training data in terms of acquisition parameters, vendors, signal-to-noise ratio (SNR), etc., the following experiments could test MoDL-QSM's robustness to these interference factors and further illustrate MoDL-QSM's ability to learn the regularization term in the physical model. Besides, the predicted $\chi_{33}$ maps were also compared with conventional and deep learning-based methods, e.g., Thresholded K-space Division (TKD) (Shmueli et al., 2009), STAR-QSM (Wei et al., 2015), AutoQSM (Wei et al., 2019a), and QSMnet (Yoon et al., 2018). The threshold in TKD method was 0.2 as suggested in the original paper. AutoQSM and QSMnet presented here were both retrained using $\chi_{33}$ as the label calculated from our training data.

**Experiment 2:** 35 scans from another two healthy volunteers (18 head orientations for the first volunteer and 17 head orientations for the second volunteer. See Table s2 for the detailed head rotation degrees) were acquired using the same scanner and imaging parameters as the training data. Three quantitative metrics: root mean squared error (RMSE), structure similarity index (SSIM) (Zhou et al., 2004), and high-frequency error norm (HFEN) (Ravishankar and Bresler, 2011) were computed on these 35 scans and compared between different methods. To further demonstrate the quantitative accuracy of MoDL-QSM in Deep Gray Matters (DGM). Five regions of interests (ROIs), putamen (PUT), globus pallidus (GP), caudate nucleus(CN), red nucleus (RN), and substantia nigra (SN) were segmented by registering a QSM Atlas (Zhang et al., 2018) to the 35 individual subjects. For each ROI, the mean and standard deviation of susceptibility reconstructed by different methods were displayed. To illustrate MoDL-QSM's ability to facilitate the physical model of single-orientation QSM reconstruction, we compared the differences between the forward-simulated phase generated by susceptibility maps of different methods with the acquired phase. The output of AutoQSM (only $\chi_{33}$), QSMnet (only $\chi_{33}$), LPCNN (COSMOS) and MoDL-QSM ($\chi_{33} + \delta B'$) were used to calculate the simulated phase maps based on their corresponding forward models, respectively. Then difference maps between the simulated phase maps and the acquired single-orientation phase maps were calculated. Mean and standard deviation of L1 errors between the simulated phase maps and the acquired single-orientation phase maps from the 35 scans were reported.

**Experiment 3:** 2016 QSM Challenge data was also used to test the reconstruction performance of AutoQSM, QSMnet, and MoDL-QSM. The reference image $\chi_{33}$ was used for ground truth susceptibility as proposed in the original paper (Langkammer et al., 2018). For quantitative comparison, RMSE, SSIM, and HFEN for the three methods were calculated.

**Experiment 4:** To explore the potential clinical applications of MoDL-QSM, pathological data including hemorrhage, multiple sclerosis (MS), and micro bleeding were used for testing. The data acquisition experiments were approved by Shanghai Jiao Tong University Human Ethics Committee and all subjects signed informed consent before scanning. The hemorrhage data were acquired using a 3D GRE sequence on a 3T GE HDxt MR scanner at Shanghai Ruijin Hospital with the following parameters: matrix size=256 × 256 × 66, voxel size=0.86 × 0.86 × 2 mm$^3$, TR=42.58 ms, TE$_1$/spacing/TE$_{16}$=3.2/2.4/39.5 ms, bandwidth=488.28 Hz/pixel, flip angle=12°. The patients with MS were acquired using a 3D GRE sequence on a 3T GE HDxt MR scanner at Shanghai Renjin Hospital with the following parameters: matrix size=256×256×124, voxel size=1 × 1 × 1 mm$^3$, TR=32.36 ms, TE$_1$/spacing/TE$_{12}$=3.2/2.4/29.2 ms, bandwidth=488.28 Hz/pixel, flip angle=12°. The micro bleeding data were acquired using a 3D GRE sequence on a 3T Philips MR scanner at Shanghai Ruijin Hospital with the following parameters: matrix size=256 × 256 × 136, voxel size=0.86×0.86×1 mm$^3$, TR=45 ms, TE$_1$/spacing/TE$_{16}$=3.3/2.5/41.9 ms, bandwidth=541 Hz/pixel, flip angle=12°.

**Experiment 5:** To test MoDL-QSM's ability for susceptibility reproducibility across different sites. Ten traveling healthy volunteers were scanned at four different sites (Shanghai Huashan Hospital, Institute of Brain and Intelligence Technology, Shanghai Jiao Tong University, and The Second Affiliated Hospital of Zhejiang University) on four 3T scanners (UIH uMR790, Shanghai, China) using a 3D GRE sequence with the following parameters: matrix size=318 × 336 × 74, voxel size=0.65 × 0.65 × 2 mm$^3$, TR=34.6 ms, TE$_1$/spacing/TE$_8$=3.3/3.7/29.2 ms, bandwidth=280 Hz/pixel, flip angle=15°. The data acquisition experiments were approved by Shanghai Jiao Tong University Human Ethics Committee and all subjects signed informed consent before scanning. The susceptibility maps were reconstructed by STAR-QSM and MoDL-QSM, respectively. Quantitatively, three DGMs: PUT, GP, and CN were segmented by registering a QSM atlas to each individual. The intraclass correlation coefficient (ICC) for repeated measurements was calculated using the SPSS (IBM Corp. Released 2012. IBM SPSS Statistics for Windows, Version 21.0. Armonk, NY: IBM Corp). A higher ICC indicates that the QSM reconstruction method can produce more similar susceptibility across the four sites. ICC in the range of 0 to 0.2 is considered to be slight, 0.2 to 0.4 is fair, 0.4 to 0.6 is moderate, 0.6 to 0.8 is substantial, 0.8-1.0 is almost perfect (Landis and Koch, 1977; Zuo and Xing, 2014).

**Experiment 6:** This experiment aims to test MoDL-QSM's performance on the data acquired at different field strengths. Ten healthy volunteers were scanned using a 3D Fast Low Angle SHot (FLASH) sequence on a whole-body 7T scanner at Fudan University (Terra; Siemens Healthineers, Erlangen, Germany) equipped with a 32-channel head coil with the following parameters: matrix size=308×358×160, voxel size=0.6×0.6×1 mm$^3$, TR=28 ms, TE$_1$/spacing/TE$_5$=4/5/24 ms, bandwidth=305 Hz/pixel, flip angle=15°. The data acquisition experiments were approved by Shanghai Jiao Tong University Human Ethics Committee and all subjects signed informed consent before scanning. Susceptibility maps were reconstructed by STAR-QSM, AutoQSM, QSMnet, and MoDL-QSM, respectively. To compare the delineation of small deep gray nuclei from surroundings on the QSM images reconstructed by different methods, the regions containing STN and GPi were zoomed in and the susceptibility profiles from STN to SN were plotted.

## 4. Results

*4.1 Comparison of susceptibility accuracy produced by COSMOS- and $\chi_{33}$- labeled networks*

Fig. 2 compares COSMOS, COSMOS labeled LPCNN output, $\chi_{33}$, and $\chi_{33}$ labeled

MoDL-QSM output on one healthy subject at four representative head orientations. Fig. 2(a) shows a representative axial slice with a manually selected ROI marked by blue. Fig. 2(b) are zoomed-in images outlined by the white box as shown in Fig. 2(a). Visually, the white matter fibers indicated by yellow arrows show similar contrast between orientations on COSMOS maps. However, the QSM maps predicted by COSMOS labeled LPCNN exhibit a clear difference between Orientation 2 and Orientation 3, indicating a clear mismatch with the labels. In contrast, MoDL-QSM can effectively preserve this underlying white matter (WM) susceptibility anisotropy, consistent with that in the labels $\chi_{33}$, as pointed by red arrows. For quantitative comparison, the WM fiber was selected as an ROI (highlighted by blue in Fig. 2(a)) to calculate the mean susceptibility values. As shown in Fig. 2(c), COSMOS shows identical susceptibility across four orientations. In contrast, COSMOS labeled LPCNN output, $\chi_{33}$, and $\chi_{33}$ labeled MoDL-QSM output exhibit susceptibility variations between orientations. Except for COSMOS, the other three results showed significant differences in susceptibility between Orientation 2 and Orientation 3 ($P$ <0.05). Additionally, there were significant differences between COSMOS and COSMOS labeled LPCNN output ($P$=0.003 for Orientation 1, $P$ =0.003 for Orientation 2, and $P$ =0.000 for Orientation 3). However, no statistically significant differences were found between $\chi_{33}$ and $\chi_{33}$ labeled MoDL-QSM output for four orientations ($P$ =0.12 for Orientation 1, $P$ =0.08 for Orientation 2, $P$ =0.54 for Orientation 3, $P$ =0.84 for Orientation 4), demonstrating the susceptibility anisotropy can be well preserved in $\chi_{33}$ labeled MoDL-QSM in brain WM.

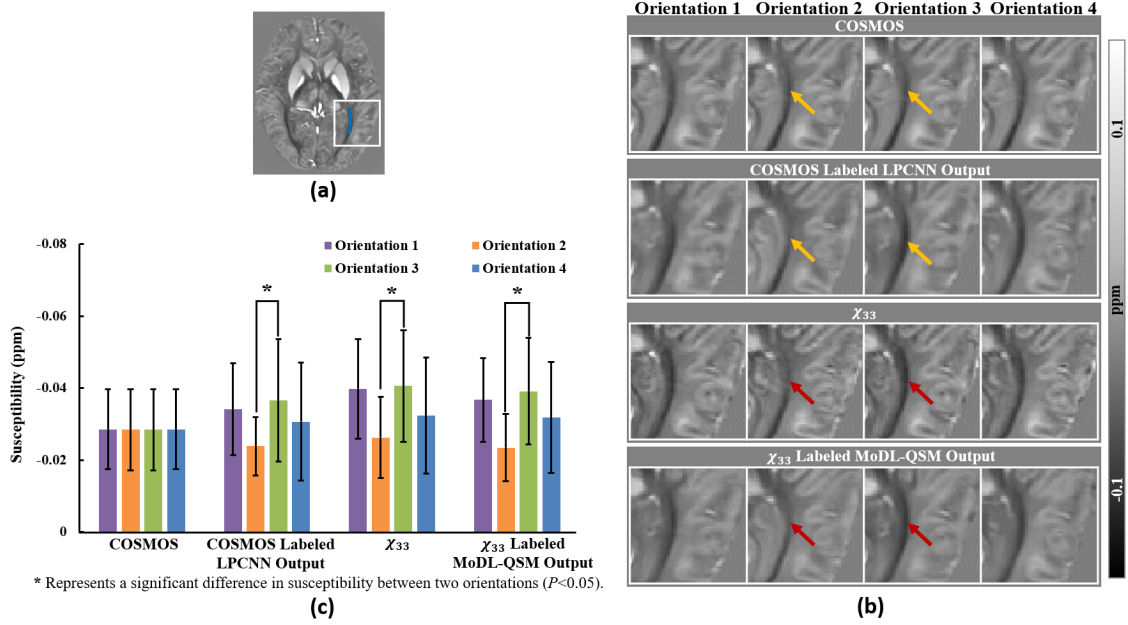

Figure 2. Comparisons of COSMOS, COSMOS labeled network output, $\chi_{33}$, and $\chi_{33}$ labeled network output on the same subject at four different head orientations. (a) A representative axial slice. (b) Zoomed-in images outlined by the white box in (a). Yellow arrows point to the white matter fiber bundles that show similar contrast in COSMOS while showing different contrast in COSMOS labeled network output at Orientation 2 and Orientation 3. Red arrows point to the preserved susceptibility anisotropy of white matter fiber bundles in both $\chi_{33}$ and $\chi_{33}$ labeled network output. (c) COSMOS shows identical susceptibility at four orientations. In contrast, COSMOS labeled LPCNN output, $\chi_{33}$, and $\chi_{33}$ labeled MoDL-QSM output exhibit consistent susceptibility changes. Except for COSMOS, the other three results show significant differences in susceptibility between Orientation 2 and Orientation 3 ($P <0.05$)

## 4.2 Evaluation of MoDL-QSM's performance

Fig. 3(a) shows three orthogonal views of QSM images on one healthy subject using five QSM reconstruction methods. Fig. 3(b) shows the zoomed-in images outlined by the red box in Fig. 3(a). The small cortical gray matter structure is more clearly observed on the QSM maps of MoDL-QSM results and label, as pointed by the yellow arrows. Fig. 3(c) displays the difference maps between the QSM maps reconstructed from different methods with respect to the label $\chi_{33}$. AutoQSM shows relatively larger differences in the cortex (the green arrow) and QSMnet shows relatively larger differences primarily in DGMs (the blue arrow). The quantitative metrics: RMSE, SSIM, HFEN of the five reconstruction methods are summarized in Table 1. The results of MoDL-QSM achieved the lowest RMSE with 54.86, the highest SSIM with 0.8801, and the lowest HFEN with 54.31, suggesting the best performances based on these criteria.

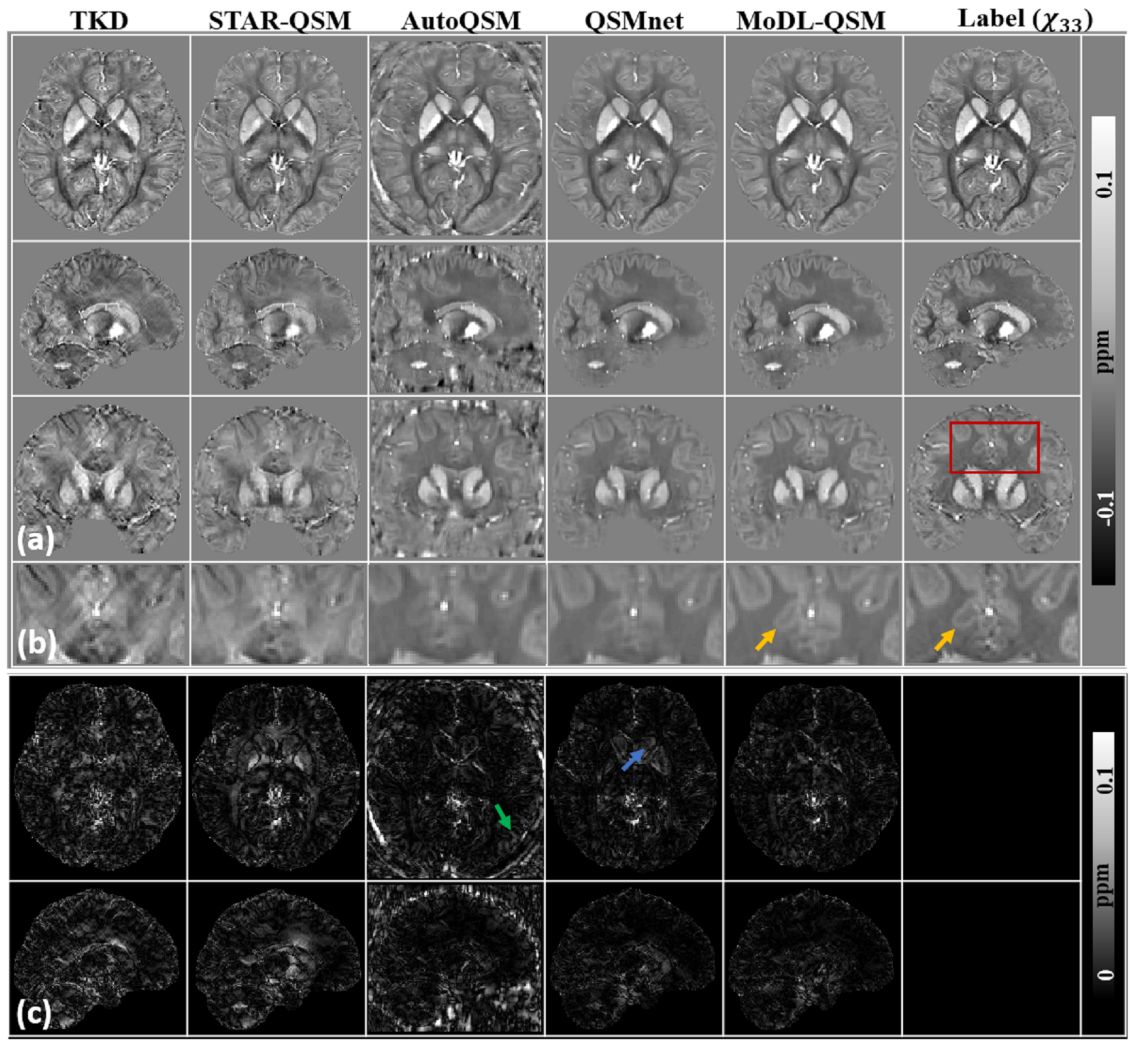

Figure 3. Comparison of different QSM reconstruction methods on a healthy subject. (a) The results are displayed in three orthogonal views. (b) Zoomed-in images of the region outlined by the red box in (a). (c) The difference maps between different reconstruction results and $\chi_{33}$ labels. Green arrow points to the larger differences in AutoQSM maps. Blue arrow points to the predicted errors in the deep gray matter on QSMnet maps.

|  | RMSE | SSIM | HFEN |
| --- | --- | --- | --- |
| TKD | 92.42±3.99 | 0.7753±0.0165 | 90.23±3.65 |
| STAR-QSM | 79.57±3.27 | 0.8180±0.0160 | 77.26±2.81 |
| AutoQSM | 69.40±2.68 | 0.8549±0.0193 | 70.04±2.73 |
| QSMnet | 55.07±3.24 | 0.8759±0.0141 | 54.39±2.70 |
| **MoDL-QSM** | **54.86±3.10** | **0.8801±0.0140** | **54.31±2.87** |

Table 1. Quantitative performance metrics, RMSE, SSIM, and HFEN from the five different QSM reconstruction methods referenced to the label $\chi_{33}$. MoDL-QSM shows better performances in all criteria than other methods.

Fig. 4 compares the regional susceptibility values in selected DGMs computed by different QSM reconstruction methods. Compared with the labels, STAR-QSM and QSMnet show underestimated susceptibility values in DGMs. In contrast, AutoQSM and MoDL-QSM have better predictions for susceptibility estimations.

Fig. 5 illustrates the reconstruction results using AutoQSM, QSMnet, and MoDL-QSM on the 2016 QSM challenge data. The results of AutoQSM and QSMnet have larger errors in GP than MoDL-QSM's results (Mean square errors in GP: 0.0097 for AutoQSM, 0.0046 for QSMnet, and 0.0026 for MoDL-QSM). RMSE, SSIM, and HFEN from the three different QSM reconstruction methods are summarized in Table 2. MoDL-QSM performs lowest RMSE with 67.91, highest SSIM with 0.8446, indicating better reconstruction results than the other two methods.

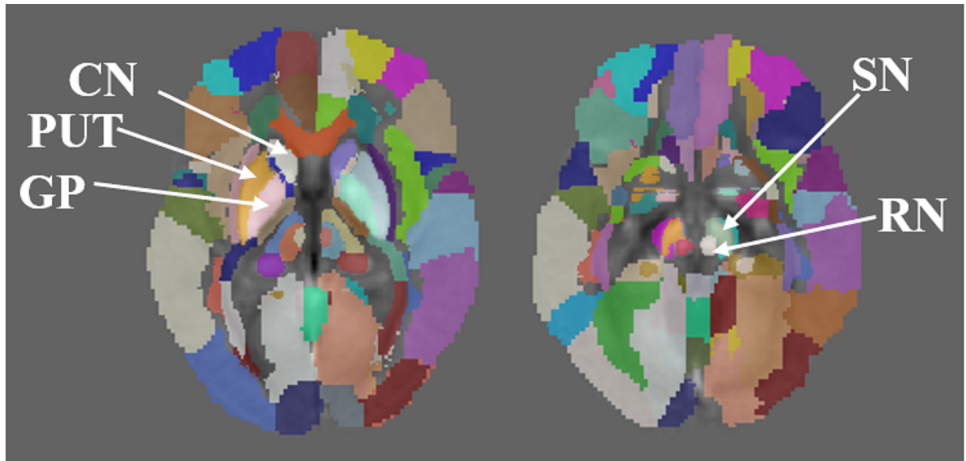

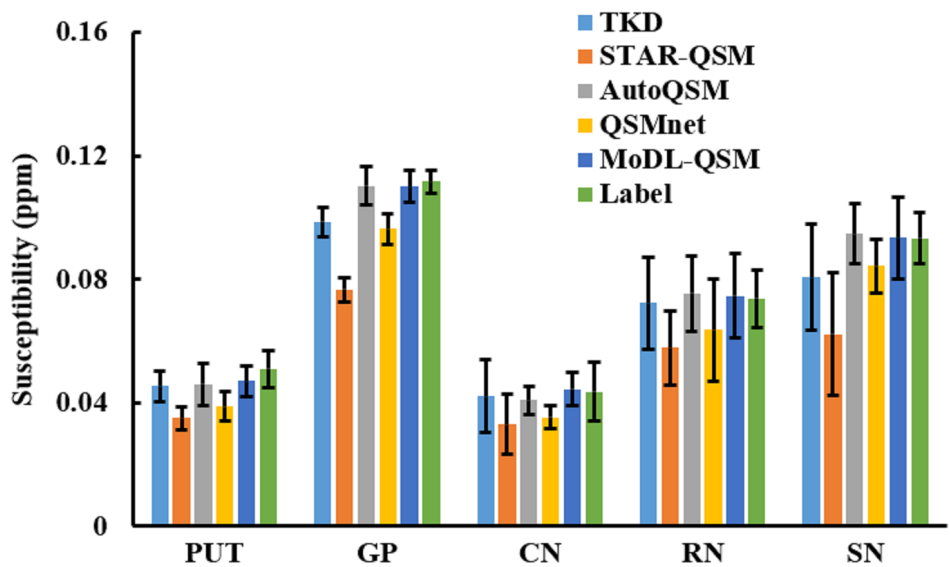

Figure 4. ROI analysis from 35 field maps of different QSM reconstruction methods. (a) QSM atlas used for ROI segmentation. (b) The susceptibility values in the ROIs (PUT, GP, CN, SN, RN). Data are presented as mean±standard deviation. The MoDL-QSM's results match well with the label.

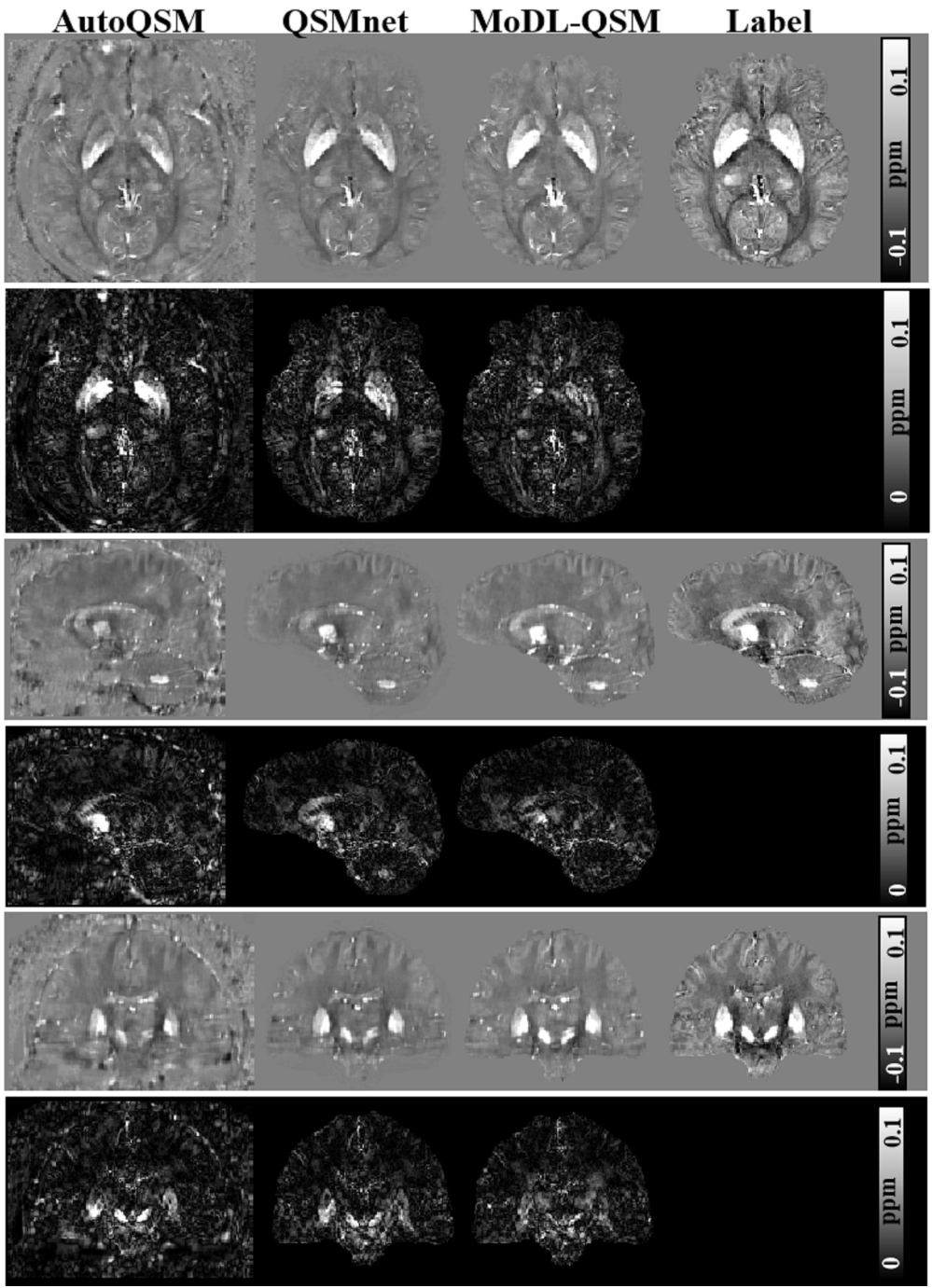

Figure 5. Comparison of different QSM reconstruction methods on the 2016 QSM Challenge data. The error map of MoDL-QSM shows a relatively smaller difference relative to the label than the compared methods.

|  | RMSE | SSIM | HFEN |
| --- | --- | --- | --- |
| AutoQSM | 78.99 | 0.8282 | 73.83 |
| QSMnet | 71.06 | 0.8386 | **66.93** |
| **MoDL-QSM** | **67.91** | **0.8446** | 67.04 |

Table 2. Quantitative performance metrics, RMSE, SSIM, and HFEN on the 2016 QSM challenge data using AutoQSM, QSMnet, and MoDL-QSM. MoDL-QSM shows better RMSE and SSIM scores than other QSM methods and a relatively higher HFEN than QSMnet.

## 4.3 Effectiveness of $\delta B'$ reconstruction

Fig. 6 displays the predicted $\delta B'$ by MoDL-QSM and the corresponding labels on one healthy subject. The results from MoDL-QSM show similar but smoother contrast compared with the label. The predicted results by MoDL-QSM reveal comparable contrasts to those observed in the label. As shown in the difference map (Fig. 6(c)), there are negligible differences related to gray and white matter. The major discrepancies that are found on the difference maps may be associated with flow effects in this non-flow compensated acquisition.

Fig. 7 compares the inconsistency between the acquired single-orientation phase and the forward-simulated field (i.e., the phase data) from the predicted susceptibility. Fig. 7(a) shows the simulated phase maps calculated from results of AutoQSM, QSMnet, LPCNN, and MoDL-QSM. The difference maps between the simulated phases and the acquired single-orientation phase are shown in Fig. 7(b). The clear differences between gray and white matter as well as within the ventricle were observed on the results of AutoQSM, QSMnet, and LPCNN. In contrast, MoDL-QSM based phase error map contains no large-scale anatomical features. L1 errors ($\times 10^{-3}$) between the simulated phase maps and the acquired single-orientation phase maps from 35 test data on two subjects were reported at the bottom of Fig. 7(b), MoDL-QSM results-based phase maps achieved the lowest error (2.88±0.09×$10^{-3}$, 3.30±0.13×$10^{-3}$, 3.93±0.16×$10^{-3}$, and 4.05±0.16×$10^{-3}$ for MoDL-QSM, LPCNN, QSMnet, and AutoQSM, respectively.).

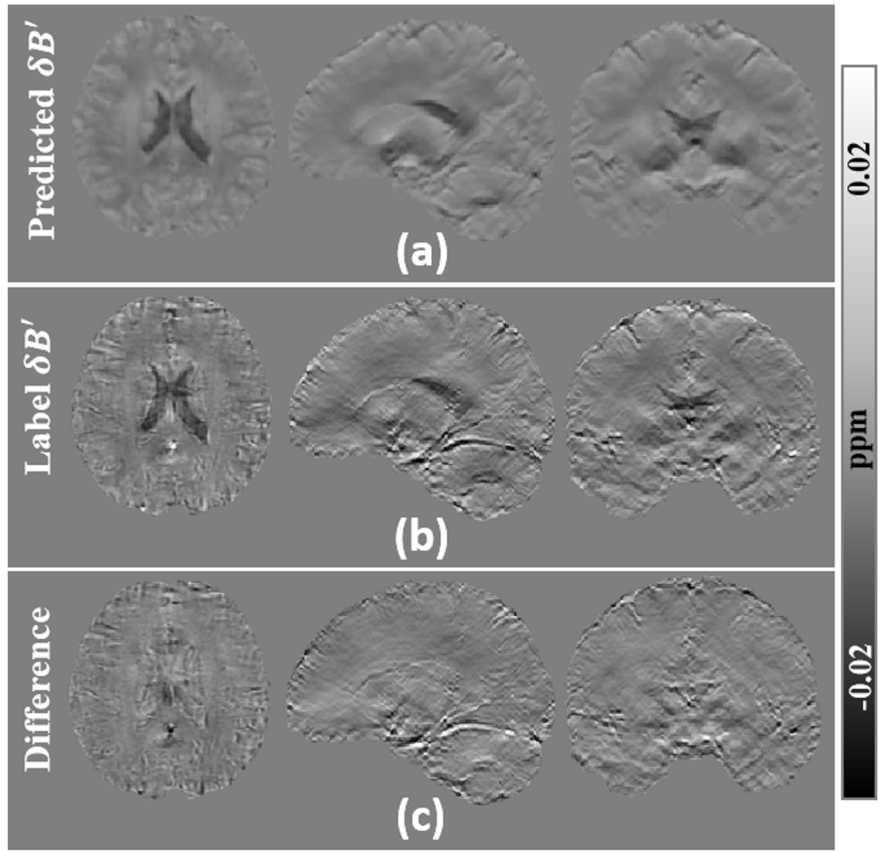

Figure 6. Comparison between predicted $\delta B'$ and label. (a) The predicted $\delta B'$ by MoDL-QSM. (b) The corresponding labels. (c) Difference maps.

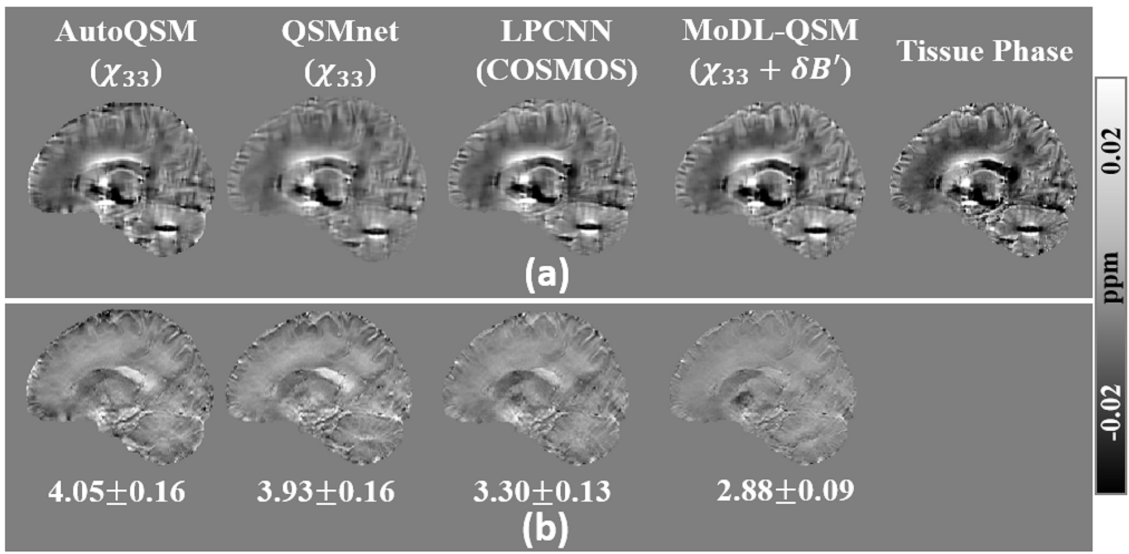

Figure 7. (a) The forward-simulated phases calculated using outputs of different networks. (b) The difference maps of forward-simulated phase from AutoQSM, QSMnet, LPCNN, and MoDL-QSM with respect to the single-orientation acquired phase. L1 errors ($\times 10^{-3}$) of each network are reported under the images.

*4.4 The practical applications of MoDL-QSM*

MoDL-QSM was also tested on patients with hematoma, multiple sclerosis (MS), and micro-bleeding, which were not used in the training dataset. The result of STAR-QSM exhibits residual blooming artifacts around the hemorrhage lesion as indicated by the blue arrow, while AutoQSM, QSMnet, and MoDL-QSM can well suppress these shadow artifacts (Fig. 8). For MS data, all QSM reconstruction methods can detect the lesion regions, and small lesions can be better visible on the results of MoDL-QSM as indicated by blue and purple arrows (Fig. 9). When applied to the patient with micro-bleeding (Fig. 10), the lesion area is similarly delineated from surroundings by all the methods as pointed by red arrows.

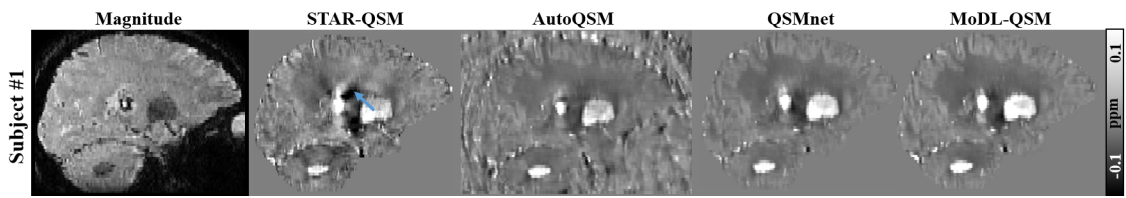

Figure 8. Representative sagittal slices of QSM images computed using STAR-QSM, AutoQSM, QSMnet, and MoDL-QSM on a patient with hemorrhage. The QSM image of STAR-QSM exhibits blooming artifacts (blue arrow) around the lesion, while AutoQSM, QSMnet, and MoDL-QSM can well suppress these shadow artifacts.

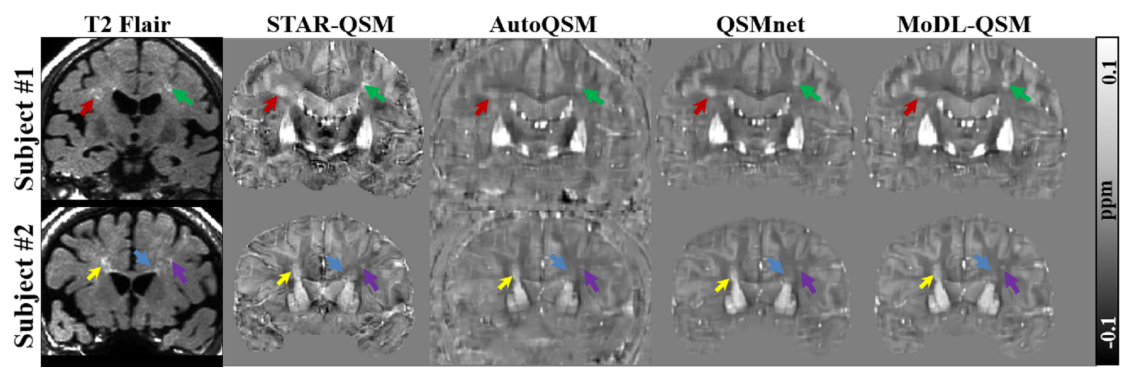

Figure 9. Representative coronal slices of QSM images computed using STAR-QSM, AutoQSM, QSMnet, and MoDL-QSM on two MS patients. The arrows point to the MS lesions.

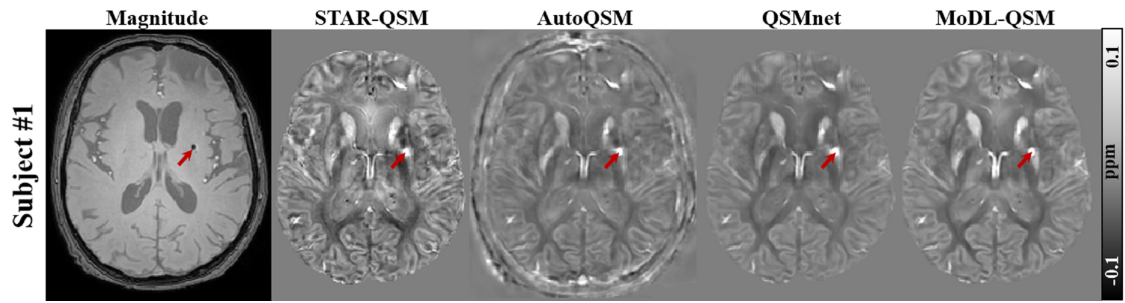

Figure 10. Representative axial slices of QSM images computed using STAR-QSM, AutoQSM, QSMnet, and MoDL-QSM on a patient with micro bleeding. The lesion area (red arrow) is similarly delineated in all the methods.

Fig. 11 shows susceptibility maps from one representative traveling subject scanned on four sites reconstructed by STAR-QSM and MoDL-QSM, respectively. The susceptibility values within the CN vary between different sites on STAR-QSM maps, as pointed by the red arrows. In contrast, MoDL-QSM results reveal more consistent susceptibility contrast in CN among different sites. Quantitatively, regional ICC of susceptibility values from four repeated scans was reported in Table 3. The susceptibility maps reconstructed by MoDL-QSM reveal higher ICC than STAR-QSM in three selected DGMs, suggesting that MoDL-QSM has better reproducibility in DGMs for the same subjects in different scanning conditions.

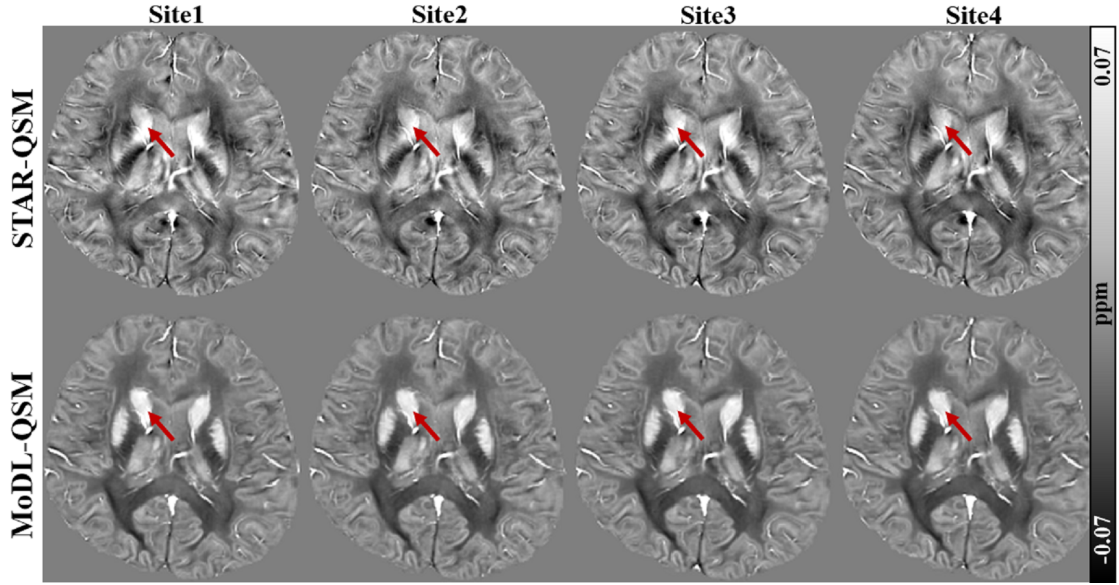

Figure 11. Susceptibility maps from a traveling subject scanned at four sites reconstructed by STAR-QSM and MoDL-QSM. Red arrows indicate the susceptibility values within the caudate nucleus vary between sites on STAR-QSM maps. In contrast, MoDL-QSM results reveal more consistent susceptibility contrast.

|  | GP | PUT | CN |
| --- | --- | --- | --- |
|  | ICC (Right/Left) | | |
| STAR-QSM | 0.700/0.819 | 0.904/0.893 | 0.731/0.639 |
| **MoDL-QSM** | **0.932/0.902** | **0.926/0.904** | **0.946/0.896** |

Table 3. The regional ICC of susceptibility values from four repeated scans. The susceptibility maps reconstructed by MoDL-QSM reveal higher ICC than STAR-QSM in both left and right sides of three selected ROIs.

Fig. 12 shows the reconstruction results on the 7T data using STAR-QSM, AutoQSM, QSMnet, and MoDL-QSM. As shown in the zoomed-in images, the STN, SN, GPi, and GPe can be observed on the QSM maps of the four methods. Fig. 12(b) plots the susceptibility profiles along the green line crossing STN and SN in Fig. 12(a), the susceptibility changes along the profile in STAR-QSM and MoDL-QSM images are relatively sharper than that in AutoQSM and QSMnet, indicating STN and SN can be better distinguished by these two methods. It should be noted that STAR-QSM suffers from artifacts around STN and SN.

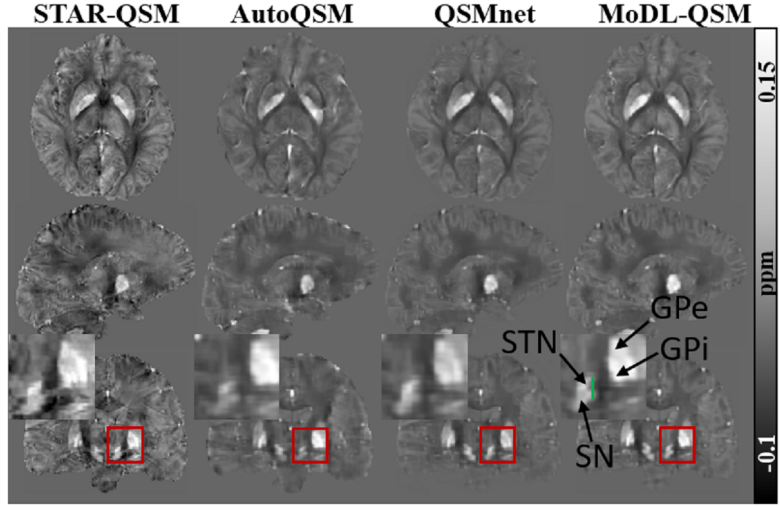

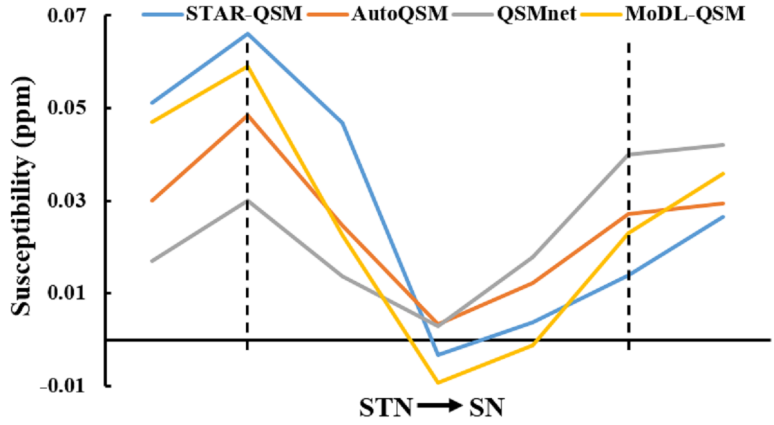

Figure 12. One representative 7T data reconstructed by STAR-QSM, AutoQSM, QSMnet, and MoDL-QSM. (a) Three orthogonal views and a zoomed-in region containing STN and GPi are shown from the region outlined by the red box. (b) Susceptibility profiles along the green line crossing STN and SN in Fig. 12(a), the susceptibility changes in STAR-QSM and MoDL-QSM are relatively sharper than AutoQSM and QSMnet.

## 5. Discussion

In this study, we proposed a model-based deep learning framework for QSM reconstruction. We incorporated the susceptibility tensor components into our proposed scheme for network training and the results showed several advantages over the state-of-the-art QSM methods: (1) the proposed deep-learning scheme accounts for the relationship between STI-derived phase induced by the rightmost column of the susceptibility tensor ($\chi_{13}$, $\chi_{23}$ and $\chi_{33}$) and the acquired single-orientation phase. (2) STI component, $\chi_{33}$, was used as the training label to make the network preserve the nature of anisotropic magnetic susceptibility in brain WM. (3) MoDL-QSM can simultaneously produce $\chi_{33}$ and the field perturbation induced by $\chi_{13}$ and $\chi_{23}$ terms. Our results showed that MoDL-QSM provided a more proper ground truth susceptibility for single-orientation QSM reconstruction. Testing on the healthy and diseased brain data demonstrates the superiority of MoDL-QSM over the compared methods.

Most of the current implementations of deep learning QSM utilize COSMOS as the ground truth susceptibility. COSMOS models susceptibility as an isotropic property, ignoring the anisotropic nature of susceptibility in WM. Thus, COSMOS susceptibility maps would not provide an accurate reference for single-orientation QSM. To mitigate this orientation bias in WM, we chose susceptibility tensor elements as the label for training our network. As shown in Fig. 2, the observed susceptibility contrast in COSMOS labeled LPCNN output is not consistent with the COSMOS label at each orientation. This is mainly because that COSMOS susceptibility reflects the effective magnetic susceptibility by averaging dipole kernels of different orientations. For model-based deep learning, the network is embedded into the physical model to learn a regularization term. In the QSM reconstruction physical model, only dipole kernel from one orientation is available. Thus, the network output preserves susceptibility anisotropy at that orientation as illustrated in Fig. 2. Unlike COSMOS, STI uses a second-order symmetric tensor to characterize the susceptibility anisotropy, where $\chi_{33}$ represents the apparent susceptibility along the z-axis. In this study, $\chi_{33}$ in the laboratory frame of reference was considered as the ground truth susceptibility for single-orientation QSM reconstruction. The consistent susceptibility changes in $\chi_{33}$ and MoDL-QSM's output (Fig. 2(c)) further demonstrate that $\chi_{33}$ can mitigate the susceptibility bias in brain WM. Therefore, we collected data from a wide range of head orientations to reconstruct high-quality STI maps and used $\chi_{33}$ as the label for network training.

Taking the tensor components of $\chi_{13}$ and $\chi_{23}$ into account, the proposed deep-learning scheme can improve the network's performance. Traditional QSM reconstruction methods ignore the contributions of $\chi_{13}$ and $\chi_{23}$ terms to the measured phase, and consider the tissue phase completely from $\chi_{33}$. In this study, we used CNN to learn the regularization term that can simultaneously reconstruct $\chi_{33}$ and the field perturbation $\delta B'$ induced by $\chi_{13}$ and $\chi_{23}$. Therefore, the framework of MoDL-QSM was more consistent with STI physical model. Additionally, quantitative performance metrics (Table 1) and ROI analysis (Fig. 4) demonstrated that MoDL-QSM benefits the incorporation of $\delta B'$ into the proposed architecture. The results of $\delta B'$ from MoDL-QSM show smoother contrast compared with the label (Fig. 6). One possible reason is that the quality of label $\delta B'$ map vary significantly between orientations due to the artifacts in $\chi_{13}$ and $\chi_{23}$, while network training acts as a regression process. In practice, the contribution of $\delta B'$ to the measured phase mostly comes from microstructure effects and chemical shielding, which may have orientation dependency. Other potential sources of errors such as vascular flux and motion might also be considered. Despite the imperfection, the forward-simulated phase from MoDL-QSM is more consistent with the acquired phase, demonstrating the phase contribution from $\delta B'$ is nonnegligible.

In MoDL-QSM, the proximal gradient descent process was unrolled into 3 iterations. Different iteration numbers were also investigated and the quantitative performance metrics of $\chi_{33}$ and $\delta B'$ were reported in Table s3 and Table s4 (see supplementary material). The reconstruction performance was improved as we increased the number of iterations from 2 to 3. However, keeping increasing the iteration number from 3 to 4 did not lead to a sustained improvement for both $\chi_{33}$ and $\delta B'$. This might be because MoDL-QSM shared weights across iterations and increasing iterative processes makes it more complicated to learn a common regularization term between the inputs and outputs in each iteration. In the future, more efforts are needed to compare the reconstruction performance in the weight shared MoDL-QSM and non-shared MoDL-QSM.

On the 2016 QSM challenge data, MoDL-QSM shows slightly higher HFEN compared with QSMnet (Table 2). The 2016 QSM challenge data was acquired using a simultaneous multiple slice (SMS) sequence, which has a lower SNR compared to our training data acquired using a 3D GRE sequence. Please note that we are ignoring noise effects in the MoDL-QSM estimations as sources of errors. The regularizer learned by MoDL-QSM may degrade performance when processing the input phases with much higher noise levels. QSMnet shows a slightly lower HFEN than MoDL-QSM when testing on this challenge data. This is likely because QSMnet contains a much larger number of parameters than MoDL-QSM, returning more resistance to noise. However, MoDL-QSM achieves better performance regarding RMSE and SSIM, demonstrating the superiority of MoDL-QSM to other compared methods.

The results of hemorrhage data showed that MoDL-QSM could well delineate lesion boundaries and suppress artifacts around large susceptibility sources (Fig. 8). The MS lesion boundary was clearly visible on MoDL-QSM images but showed ambiguity on STAR-QSM images (Fig. 9), suggesting the potential of MoDL-QSM in detecting and characterizing small MS lesions in clinical applications.

Furthermore, the multicenter traveling subject experiment was also carried out in this study. In the repeated scans at four sites, the traveling subjects were scanned in the normal supine position. Therefore, the head orientation related to the main magnetic field could be similar between scans. Additionally, DGMs with mainly iron deposits are considered to be isotropic tissues. Therefore, the susceptibility in DGMs should be identical between different scans. As shown in Fig. 11, the susceptibility of the caudate nucleus in STAR-QSM maps varies between scans while that in MoDL-QSM maps are more consistent. The higher ICC values across different sites in three DGMs of MoDL-QSM further support these observations quantitatively. The results indicate that MoDL-QSM is more robust to the interferences, e.g., the distance between head position and coil, thermal noise level when the subjects were scanned longitudinally. The advantages of high reproducibility and less quantification error demonstrate MoDL-QSM's potential for reliable longitudinal measurement of susceptibility time courses, enabling more precise monitoring for metal ions accumulation in neurodegenerative disorders, e.g., Parkinson's disease and Alzheimer's disease.

The content of cortical iron has been increasingly recognized as a biomarker in Alzheimer's disease and cognitive decline (Bulk et al., 2018; Chen et al., 2020a; Damulina et al., 2020; van Duijn et al., 2017). However, the artifacts between cortical gray and white matter provided by conventional reconstruction methods hinder QSM's precision in cortical iron quantification. In this study, the results of MoDL-QSM showed that artifacts in the cortical areas could be well suppressed and gray-white matter boundaries can be revealed (Fig. 2&3). This outcome may provide great potential to achieve iron quantification in the high-resolution *ex vivo* cortex and in the superficial white matter (Kirilina et al., 2020), which is helpful to further investigate the pathogenesis of Alzheimer's disease.

The benefit of ultrahigh-field 7T for QSM has been recognized as increasing interest (Bian et al., 2016; Li et al., 2012). However, the MRI signal acquired at 7T is more sensitive to field inhomogeneity and suffers from more rapid decay that causes a higher noise level when high-resolution data is required. These factors bring challenges to the dipole inversion problem for high-quality QSM. The resulting images suffer from poor quality and streaking artifacts around the region with large susceptibility variations. As illustrated in Fig. 12, the STAR-QSM results show artifacts around STN and GPi. While these artifacts on MoDL-QSM maps are invisible, suggesting MoDL-QSM can provide confident tissue structural boundaries. The improved visualization makes it possible for 7T QSM to guide DBS, which is valuable since the accurate electrode localization has been proven critical for a successful outcome, such as for STN and GPi DBS for PD patients (Kelman et al., 2010).

The trained MoDL-QSM has some limitations. Firstly, MoDL-QSM takes both $\chi_{33}$ and $\delta B'$ into account to make the proposed framework more consistent with STI physical model. However, the tissue phase originates not only from magnetic susceptibility but also from the chemical shift, chemical exchange, and complex tissue microstructure. These sources are not considered in the STI model or might be incorrectly assigned to STI components. Secondly, in Fig. 7, we demonstrated $\delta B'$ is a part of field perturbation that could not be ignored. The potential sources and applications of $\delta B'$ need to be further investigated.

## 6. Conclusion

We proposed an STI model-based deep learning framework for single-orientation QSM reconstruction. We first demonstrated that $\chi_{33}$ is more rational as the label than COSMOS. Thanks to the powerful feature extraction and characterization ability of CNN, we incorporate both $\chi_{33}$ and phase contributions from $\chi_{13}$ and $\chi_{23}$ to force MoDL-QSM to follow the STI model. Qualitative and quantitative comparisons show that MoDL-QSM has superior reconstruction performance than compared QSM methods. MoDL-QSM offers a way to generate high-quality QSM and may provide clinical values for well characterizing the lesions, e.g., MS or providing high fidelity anatomical delineation of brain nucleus for DBS targeting.

**Acknowledgments**

This study is supported by the National Natural Science Foundation of China (61901256, 91949120, 62071299).

619-626.

De Rochefort, L., Brown, R., Prince, M.R., Wang, Y., 2008. Quantitative MR susceptibility mapping using piece‐wise constant regularized inversion of the magnetic field. Magnetic Resonance in Medicine: An Official Journal of the International Society for Magnetic Resonance in Medicine 60, 1003-1009.

Du, L., Zhao, Z., Cui, A., Zhu, Y., Zhang, L., Liu, J., Shi, S., Fu, C., Han, X., Gao, W., 2018. Increased iron deposition on brain quantitative susceptibility mapping correlates with decreased cognitive function in Alzheimer's disease. ACS chemical neuroscience 9, 1849-1857.

Gao, Y., Zhu, X., Moffat, B.A., Glarin, R., Wilman, A.H., Pike, G.B., Crozier, S., Liu, F., Sun, H., 2020. xQSM: quantitative susceptibility mapping with octave convolutional and noise‐regularized neural networks. NMR in Biomedicine, e4461.

Goodfellow, I., Pouget-Abadie, J., Mirza, M., Xu, B., Warde-Farley, D., Ozair, S., Courville, A., Bengio, Y., 2014. Generative adversarial nets. Advances in neural information processing systems 27, 2672-2680.

Haacke, E.M., Liu, S., Buch, S., Zheng, W., Wu, D., Ye, Y., 2015. Quantitative susceptibility mapping: current status and future directions. Magnetic resonance imaging 33, 1-25.

Hammernik, K., Klatzer, T., Kobler, E., Recht, M.P., Sodickson, D.K., Pock, T., Knoll, F., 2017. Learning a variational network for reconstruction of accelerated MRI data. Magnetic Resonance in Medicine 79.

He, K., Zhang, X., Ren, S., Sun, J., 2016. Deep residual learning for image recognition. Proceedings of the IEEE conference on computer vision and pattern recognition, pp. 770-778.

Jenkinson, M., Smith, S., A global optimisation method for robust a ne registration of brain images. Medical Image Analysis 5, 143.

Jung, W., Yoon, J., Ji, S., Choi, J.Y., Kim, J.M., Nam, Y., Kim, E.Y., Lee, J., 2020. Exploring linearity of deep neural network trained QSM: QSMnet+. Neuroimage 211, 116619.

Kames, C., Doucette, J., Rauscher, A., 2019. Proximal variational networks: generalizable deep networks for solving the dipole-inversion problem. 5th International QSM Workshop.

Kelman, C., Ramakrishnan, V., Davies, A., Holloway, K., 2010. Analysis of stereotactic accuracy of the cosman-robert-wells frame and nexframe frameless systems in deep brain stimulation surgery. Stereotactic & Functional Neurosurgery 88, 288-295.

Kingma, D., Ba, J., 2015. Adam: A method for stochastic optimization in: Proceedings of the 3rd international conference for learning representations (ICLR'15). San Diego.

Kirilina, E., Helbling, S., Morawski, M., Pine, K., Reimann, K., Jankuhn, S., Dinse, J., Deistung, A., Reichenbach, J.R., Trampel, R., Geyer, S., Müller, L., Jakubowski, N., Arendt, T., Bazin, P.-L., Weiskopf, N., 2020. Superficial white matter imaging: Contrast mechanisms and whole-brain in vivo mapping. Science Advances 6, eaaz9281.

Kressler, B., De Rochefort, L., Liu, T., Spincemaille, P., Jiang, Q., Wang, Y., 2009.